# Contrastive Learning for Efficient Transaction Validation in UTXO-based Blockchains

Hamid Attar
hamid1attar@yahoo.co.uk

Luigi Lunardon
Teranode group
Zug, Switzerland
l.lunardon@teranode.group

Alessio Pagani
University of Bath
Bath, UK
ap3060@bath.ac.uk

*Abstract*— This paper introduces a Machine Learning (ML) approach for scalability of UTXO-based blockchains, such as Bitcoin. Prior approaches to UTXO set sharding struggle with distributing UTXOs effectively across validators, creating substantial communication overhead due to child-parent transaction dependencies. This overhead, which arises from the need to locate parent UTXOs, significantly hampers transaction processing speeds. Our solution uses ML to optimize not only UTXO set sharding but also the routing of incoming transactions, ensuring that transactions are directed to shards containing their parent UTXOs. At the heart of our approach is a framework that combines contrastive and unsupervised learning to create an embedding space for transaction outputs. This embedding allows the model to group transaction outputs based on spending relationships, making it possible to route transactions efficiently to the correct validation microservices. Trained on historical transaction data with triplet loss and online semi-hard negative mining, the model embeds parent-child spending patterns directly into its parameters, thus eliminating the need for costly, real-time parent transaction lookups. This significantly reduces cross-shard communication overhead, boosting throughput and scalability.

*Keywords—blockchain, scalability, bitcoin, neural networks, contrastive learning, machine learning*

I. INTRODUCTION

Blockchain technologies have transformed transaction management by introducing decentralized, secure, and transparent systems for value transfer. Unlike traditional centralized systems, blockchain eliminates the need for intermediaries, thereby increasing transparency and robustness against single points of failure. However, these benefits come with significant scalability challenges. As the number of users, transactions, and diverse use-cases increases, ensuring the integrity and security of the ledger becomes more complex. The system must handle a growing volume of transactions without compromising on performance or security, which is crucial for broader adoption. In Unspent Transaction Output (UTXO)-based blockchains like Bitcoin [1] or Cardano [2], each transaction consumes existing UTXOs and may create new ones. This UTXO model enables precise tracking of funds through the ledger, but it also poses unique challenges for scalability and efficiency as transaction volumes grow.

Fundamentally, blockchain is a decentralized digital ledger that records transactions in blocks linked sequentially, forming an immutable chain. In UTXO-based blockchains, each UTXO represents a discrete unit of value that can only be spent once, ensuring traceability and preventing double-spending. To spend a UTXO, a transaction must satisfy its locking conditions. In Bitcoin, these are set in Bitcoin Script using operational (OP) codes that define logic for verifying ownership. Locking scripts specify the conditions required for the UTXO to be spent, such as matching a cryptographic signature. Unlocking scripts, provided by the spender, must meet these conditions for the referenced UTXOs to be spent.

Transactions are validated by ensuring that the inputs of each transaction correspond to unspent outputs from previous transactions. Validators on the blockchain network check the UTXO locking conditions, verify that the referenced UTXOs exist and are unspent, and ensure that the transaction outputs are sufficiently funded by the UTXOs being spent. Typically, transactions are validated sequentially, which creates a bottleneck as transaction volumes increase. Verifying each transaction's validity one at a time adds latency and limits throughput, especially as the UTXO set grows. This process requires accessing the entire UTXO set, and, when coupled with a massive number of transactions to validate, results in significant computational and storage overhead. These compounding factors create formidable scalability challenges that hinder the blockchain's ability to effectively accommodate future growth and user demand.

Sharding has emerged as a promising solution to address these challenges by distributing transactions across different validators to enable parallel transaction processing [3], [4]. This approach enhances throughput and reduces network latency, making scalability feasible as transaction volumes increase. However, in UTXO-based blockchains, when a new transaction is received, it must be routed to the correct validator where the relevant parent UTXOs are located (i.e., in which shard). Without efficient routing, transactions would incur high communication overhead to retrieve missing UTXOs needed for validation, slowing processing times.

We present an approach to sharding by harnessing machine learning (ML) techniques. Our method poses potential to enhance the number of transactions per second a UTXO-based blockchain node can handle by partitioning the UTXO set into shards for each validator available within one node, and routing incoming transactions to the shard likely to contain the parent UTXOs, enabling parallel transaction validation at reduced communication overhead. At the core of our solution is a dual-model framework that uses contrastive and unsupervised learning to effectively group transaction outpoints, which are represented as feature vectors capturing essential attributes of each transaction. The objective is to develop a ML model that can distribute transactions to validators respecting parent-child relationships, using only the information received within the transactions without fetching additional data. By embedding these features into a structured space, the model learns relationships between transactions, such as identifying parent-child linkages. Once this embedding is established, we can cluster the transaction outpoints into distinct groups, or shards, each assigned to a specific transaction validator.

## II. RELATED WORK

Several projects have developed sharding techniques to boost blockchain scalability and transaction throughput. OmniLedger [5] partitions the network into independent shards to enhance parallelism, using the Atomix protocol for secure, atomic transaction processing across shards. Similarly, RapidChain [6] takes a committee-based approach, assigning transaction subsets to committees to scale parallelization. Initially, it proposed that all nodes store the details of every committee, which proved impractical due to data burden and security risks. Instead, RapidChain adopted a Kademlia-inspired [7] routing protocol, where each committee maintains a local routing table, enabling efficient, log-time transaction forwarding. Chainspace [8], while designed for smart contracts, introduced a sharding model adaptable to UTXO-based systems, with Byzantine Fault-Tolerant consensus within each shard and the Sharded Byzantine Atomic Commit (S-BAC) protocol for atomic cross-shard transactions, supporting scalable, parallel execution across shards.

While blockchain sharding has improved performance, high cross-shard ratios still lead to significant processing overhead. Aeolus [9] addresses this by segmenting transactions into stages for parallel execution, optimizing transaction allocation and enhancing throughput. Similarly, OptChain [10] developed a transaction placement strategy designed to lower the cross-shard ratio by grouping highly correlated transactions within the same shard. However, this approach necessitates that clients execute the transaction placement algorithm, which can burden lightweight clients and presents certain limitations. Pyramid [4] introduced a hierarchical system architecture that enables specific nodes in two related shards to store the states of both, effectively converting cross-shard transactions into intra-shard ones but at the cost of higher storage and computational demands. Similarly, Brokerchain [11] introduces broker accounts to convert cross-shard transactions into intra-shard ones, but a larger shard count can lead to excessive additional transactions. Solutions like Bitcoin SV Teranode [12] and Monoxide [13] proposed shard allocation based on transaction or address hashes, but this random distribution often leaves shards without necessary parent UTXOs, increasing communication and processing strain. Optimizing UTXO allocation remains crucial for scaling UTXO-based blockchains like Bitcoin.

Recent advances in Artificial Intelligence (AI) have opened new avenues for optimizing processes in a variety of fields, and blockchain technology is no exception. In this regard, community detection-based sharding was used to address high cross-shard ratios and performance bottlenecks. This approach leverages account-based graphs, where nodes (addresses) and edges represent spending relationships, clustering frequently interacting accounts to reduce cross-shard transactions and enhance scalability. Previous studies [14], [15] use the Metis graph partitioning algorithm for account-based sharding, which minimizes edge weights between subgraphs but doesn't capture transaction correlation, resulting in a high cross-shard ratio. Li et al. [16] improved on this with CLPA, a community detection algorithm that iteratively updates transaction relationships, reducing cross-shard ratios and balancing load across shards. However, CLPA's iterative nature incurs high latency and frequent account migrations, limiting scalability, especially with large shard numbers, reported by Wu et al. [3].

Building on the transactional account-based graphs discussed, another critical graph structure within blockchain networks focuses on the relationships between transactions themselves, where each transaction forms a node and the edges represent the flow of value between transactions. This structure is best represented as a Directed Acyclic Graph (DAG), capturing the spending relationships across blockchain transactions. Unlike the linear chain of blocks, this DAG enables a deeper view of transaction histories, as edges denote outpoints, tracing value as it moves from one transaction to the next. To learn these inter-transaction relationships effectively, Graph Neural Networks (GNNs) offer a powerful approach, efficiently encoding each transaction's place in the graph through node embeddings that reflect both structural and transactional patterns. This capability has proven particularly valuable in detecting suspicious activities, as seen in the work of Weber et al. [17] using Graph Convolutional Networks (GCNs) for anti-money laundering in Bitcoin. Similarly, Geng et al. [18] and Li et al. [19] applied GNNs to trace and predict illicit behaviors, with results suggesting that embeddings computed through GNNs enhance both the interpretability and accuracy of transaction tracking models, underscoring their relevance for scalable blockchain analysis. However, while Graph-based node embeddings capture spending relationships in blockchain transactions effectively, constructing these graphs requires frequent lookups based on parent TXIDs and outpoint indexes, adding latency and computational load from real-time blockchain queries, which limits scalability for high-throughput transaction validation.

Our goal is to eliminate these lookups during inference by embedding spending relationships directly within the model's learnt weights during training, improving efficiency and scalability. By learning a metric space in which parent UTXOs and child transactions are positioned closely together, the model can use this space for sharding decisions. Through contrastive learning, we develop a model trained using the triplet loss function [20], [21] with online hard negative triplet mining where positive pairs represent parent-child transactions and negative pairs represent unrelated transactions. At inference, our model requires only the features of the new transaction, embedding spending patterns directly into the model parameters to bypass parent transaction lookups, thus reducing communication overhead and enhancing the scalability of UTXO-based blockchains. The main novelties of our approach lie in our embedding model and encompass three key aspects:

- **Elimination of Communication with Blockchain During Model Inference:** Unlike graph-based models that require fetching parent transaction data, our method operates solely with the information immediately available when a transaction is received.

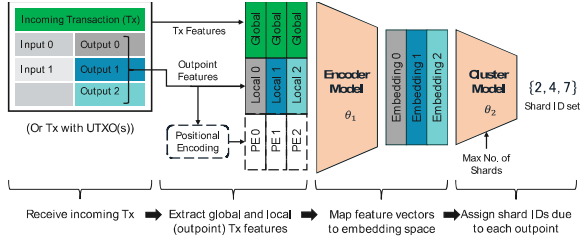

Fig. 1 Framework for mapping transactions (tx) to points in embedding space by $f_{\theta_1}$, and subsequently assigning them to shard IDs by $g_{\theta_2}$.

- **Creation of Transaction Outpoint Embeddings:** Our method ensures that embeddings of parent-child transactions or transaction outpoints are similar under a chosen metric. This allows us to use learned weights, capturing spending relationships during training, to embed new transaction outpoints consistently with their parents. A clustering model then identifies clusters from these learnt embeddings and assigns shard IDs, which determine transaction routing to microservices.

- **Training with Triplet Loss:** Our training uses a triplet loss with online hard negative triplet mining on triplets of transactions or transaction outpoints, dynamically selecting hard and semi-hard negatives within the embedding space as it evolves with training. This method ensures the model effectively captures spending relationships during training.

### III. SOLUTION METHODOLOGY

This section details our solution to achieve effective clustering of transactions with their parent UTXOs, critical for identifying shards for routing of transactions to validation microservices. Our approach hinges on a Multi-Layer Perceptron (MLP) as $f_{\theta_1}$ in Fig. 1 trained using contrastive learning. Our framework is designed to capture the spending relationships inherent in between transactions without requiring real-time parent transaction lookups.

#### A. Duel-Model Architecture for UTXO Clustering

We propose a dual-model architecture which is formed based on a combination of contrastive learning and unsupervised learning that performs spending-based clustering on transaction outpoints. Initially, feature vectors $x \in \mathbb{R}^d$ representing outpoints are fed into an encoder model $f_{\theta_1}: \mathbb{R}^d \to \mathcal{E}$, which learns to project these vectors into a structured embedding space $\mathcal{E} \subseteq \mathbb{R}^k$. The goal is to map parent and child transactions into proximity within the embedding space, even if they are far apart in the original feature space. This alignment ensures that transactions sharing a common lineage are grouped closely in the embedding space. Once a meaningful and well-organised embedding space $\mathcal{E}$ is achieved, a cluster model $g_{\theta_2}: \mathcal{E} \to \{C_1, C_2, \ldots, C_n\}$ is applied to identify clusters $C$ of these transaction outpoints. These clusters then form shards, which are distributed across the validation microservices within one node on the network. For new incoming transactions that node receives, the outpoint feature vectors $x_{\text{new}}$ are mapped into the embedding space $\mathcal{E}$ and assigned a shard ID based on their cluster $C_i$. The set of shard IDs $\{C_i\}$ associated with a transaction determines the specific microservices to which the transaction will be routed for validation. An overview of the framework described is presented in Fig. 1.

TABLE I. FEATURES FOR BITCOIN TRANSACTIONS. SIZE DENOTES NUMBER OF ELEMENTS IN THE FEATURE VECTOR. GLOBAL DENOTES A DESCRIPTOR OF THE TRANSACTION AS A WHILE LOCAL DENOTES OUTPOINT SPECIFIC.

| Size | Type | Feature | Description |
|---|---|---|---|
| 1 | Global | No. inputs | Number of transaction inputs |
| 1 | Global | No. outputs | Number of transaction outputs |
| 1 | Global | Sum outputs | Total Satoshi output amount |
| 6 | Global | Output statistics | Max, min, mean, mode, median, std of output amounts |
| 1 | Global | Tx size | Sum of length of all scripts (inputs and outputs) |
| 5 | Global | Encodings of top 5 unlocking scripts | Takes on values of 0 if the input does not exist within the top 5; 1 if it exists but is not a P2PKH unlocking script; 2 if it exists and it is a P2PKH unlocking script |
| 5 | Global | Encodings of top 5 locking scripts | Takes on values of 0 if the output does not exist within the top 5; 1 if it exists but is not a P2PKH locking script; 2 if it exists and it is a P2PKH locking script |
| 76 | Local | Locking Script | Bag of OP codes encoding (see Table III for examples) |
| 1 | Local | Locking Script Size | Length of locking script |
| 1 | Local | Amount | Satoshi amount |
| 10 | Positional | Positional Encoding | Vector for a given outpoint position within a transaction [22] |

#### B. Feature Extraction from Bitcoin Transactions

The fundamental unit in our framework is a transaction outpoint, which we transform into a three-part feature vector. We derive a novel set of features for transaction outpoints, summarized in Table I. Global features include input/output counts, statistics of output amounts, and the encodings of the unlocking and locking scripts. Local features focus on outpoint-specific attributes, such as the amount and the size of the locking script. Positional encodings [22] differentiate identical outpoints. These components are concatenated to create the feature vector, serving as the input to our model in Fig. 1. Since $f_{\theta_1}$ must efficiently shard incoming transactions in real-time, it relies solely on immediately available data when receiving an incoming transaction. This constraint limits our features to those available with the incoming transaction.

To encode the locking script information for outpoint features, we propose Bag of OP Codes - an encoding like bag of words in Natural Language Processing (NLP) but applied to Bitcoin scripts. We create a vocabulary of OP codes, exemplified in Table II. Each locking script (i.e., scriptPubKey) in a transaction output is then analyzed. For each OP code in the vocabulary, if it appears in the scriptPubKey, the corresponding vector index position is set to 1; otherwise, it is set to 0. This binary vector representation captures the presence or absence of specific OP codes within the script. Additionally, an "Other" category is included for any out-of-vocabulary OP codes, ensuring comprehensive coverage of the script's operational content. Examples of Bag of OP codes encodings for common locking scripts are shown in Table III. This method provides a structured way to encode and compare Bitcoin scripts based on their OP code composition.

TABLE II. EXAMPLE VOCABULARY FOR BAG OF OP CODES.

| Index | OP Code |
|---|---|
| 0 | OP_DUP |
| 1 | OP_HASH160 |
| 2 | OP_EQUALVERIFY |
| 3 | OP_DROP |
| 4 | OP_CHECKSIG |
| 5 | OP_CHECKLOCKTIMEVERIFY |
| 6 | Other |

## C. Training

*1) Loss Function:* We employ the triplet loss for training $f_{\theta_1}$ as introduced in the FaceNet paper [21], Triplet loss is widely used in ML and deep learning for embedding tasks, like face recognition and text similarity, aiming to bring similar samples closer and push dissimilar ones apart in the embedding space. It operates on sample triplets—an anchor, a positive (similar) example, and a negative (dissimilar) example—minimising the anchor-positive distance while maximising the anchor-negative distance. This approach enforces a margin, promoting better discrimination and clustering in the learned embedding space, which is particularly useful for distinguishing subtle differences and grouping related items. To formalize this, we define a distance metric $d(h, h')$ as the Euclidean distance between two embeddings $h$ and $h'$. The objective is to enforce that the distance between the anchor $h^a$ and positive $h^p$ is less than the distance between the anchor and negative $h^n$ by a margin $m$, which can be expressed as in (1).

$$d(h^a, h^p) + m < d(h^a, h^n) \quad (1)$$

The triplet loss is thus constructed to enforce this constraint. For a given triplet $(h^a, h^p, h^n)$, the triplet loss $\mathcal{L}$ is defined as

$$\mathcal{L} = \max(0, d(h^a, h^p) - d(h^a, h^n) + m) \quad (2)$$

where $\max(0, \cdot)$ ensures that the loss is non-negative and only contributes when the constraint is violated. By substituting the Euclidian distance into this formula, we obtain (3).

$$\mathcal{L} = \max(0, \|h^a - h^p\|_2 - \|h^a - h^n\|_2 + m) \quad (3)$$

The batch loss is calculated as the average over all valid triplets, ensuring that embeddings of anchors and positive pairs are brought closer, while anchors and negatives are pushed apart, thereby structuring the embedding space in a way that captures the desired similarities and relationships. In our case, positive pairs are parent-child transaction outpoints, and negatives are unrelated outpoints. This setup captures transactional lineage, embedding spending patterns directly into the model. Thus, at inference, only the features of the incoming transaction are needed, as spending relationships are already embedded in the model parameters.

TABLE III. EXAMPLES OF BAG OF OP CODES ENCODINGS.

| Type | scriptPubKey | Bag of OP Codes Vector |
|---|---|---|
| P2PKH | OP_DUP OP_HASH160 \<pubKeyHash\> OP_EQUALVERIFY OP_CHECKSIG | [1, 1, 1, 0, 1, 0, 0] |
| P2PK | \<pubKey\> OP_CHECKSIG | [0, 0, 0, 0, 1, 0, 0] |
| Locktime | \<expiry_time\> OP_CHECKLOCKTIMEVERIFY OP_DROP OP_DUP OP_HASH160 \<pubKeyHash\> OP_EQUALVERIFY OP_CHECKSIG | [1, 1, 1, 1, 1, 1, 0] |

*2) Selection of Negative Pairs:* Incorporating the margin mmm and the online selection of negative samples is vital for creating well-separated and generalisable embeddings. The margin defines a buffer zone that maintains separation between positive and negative pairs, ensuring significant discrimination rather than mere proximity. Negative samples challenge the model during training, enabling it to effectively distinguish between related (positive) and unrelated (negative) transaction outpoints.

In our use case, selecting positive pairs is straightforward, as they derive from connected transactions in the spending graph, specifically parent-child transaction outpoints. The selection of negative samples for a given anchor, however, is critical to training the model effectively [20], [21], [23]. A well-chosen negative sample ensures convergence and results in an organised embedding space, where parent and child transactions are in proximity. To achieve the desired embedding space, we use online hard negative sampling, which dynamically selects negative samples during training based on the current state of the embedding vectors. This approach aims to choose sufficiently challenging negative samples to enhance the model's performance while preventing it from collapsing to a single point. Three categories of negatives can be identified, illustrated in Fig. 2.

1. **Easy:** These negatives are already far from the anchor and beyond the margin $m$ in the embedding space (Fig. 2. right). Their inclusion offers minimal learning benefit, as the model can easily distinguish them from positive pairs, resulting in small gradients and stagnant learning.

2. **Hard:** These negatives are closer to the anchor than the positive (Fig. 2, left). Their exclusive use can be harmful, as they may result in trivial solutions learnt, causing model collapse (e.g., $f_{\theta_1}(x) = 0$), overwhelming it and resulting in incorrect adjustments in embedding space.

3. **Semi-Hard:** These negatives are further from the anchor than the positive but still within a challenging distance (Fig. 2, center). They provide a balance, being challenging enough to facilitate learning without causing model collapse. Training with semi-hard negatives allows for gradual refinement of the embedding space, ensuring correct clustering of parent and child transactions without overwhelming the learning process.

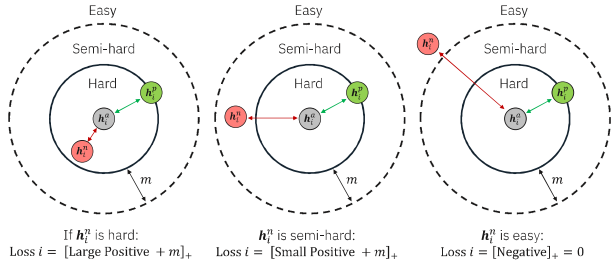

Fig. 2 Illustration of three types of negative samples: hard, semi-hard, easy.

We recommend pre-training the model on semi-hard negatives, which helps establish a robust embedding space. As training progresses and the loss function stagnates, we gradually introduce hard negatives to refine the model's ability to distinguish complex relationships. This approach maximizes accuracy achievable with hard negatives while mitigating the issues associated with their exclusive use.

*D. Dataset*

Our approach is designed to generalize across any UTXO-based blockchain. In this study, we used BSV transaction data due to its large blocks and diverse data, which demand efficient validation. This dataset included 6,000 blocks from 21/06/2024 to 01/08/2024, excluding coinbase, yielding 102,761,293 transactions (80 GB), split 80/20 for training and testing.

## IV. RESULTS

*A. Accuracy of Clustering*

We evaluated our framework (Fig. 1) for its ability to group parent-child transaction outpoints. Accuracy was measured by the probability of correctly assigning parent-child to the same shard (Fig. 2). Higher values reflect reduced cross-shard communication. Two benchmarks based on random allocation are shown: Rand One (one random shard per transaction) and Rand Many (one random shard per transaction outpoint). The dashed lines represent ours. Results show a substantial improvement over random methods, especially up to 10 shards (red vertical line), achieving >90% accuracy in shard assignment. Beyond 10 shards, accuracy declines as clusters are split, akin to hierarchical clustering; we saw ~10 distinct point clusters in the embedding space on our dataset, likely due to inherent transaction patterns rather than model imposition. When misassignment occurs, examining the K nearest neighboring shards (i.e., clusters adjacent to the predicted one) considerably enhances the probability of locating the parent UTXO. For instance, if the parent is absent from the predicted shard out of 100 shards, checking the 3 nearest shards to fetch the parent UTXO improves accuracy from 47% to 70%, while checking the 9 nearest shards boosts it to 80%. When K>0, accuracy reflects the likelihood of finding the parent within these neighboring shards, owing to a meaningful embedding space that captures spending relationships as metric distances.

*B. Quantitative Evaluation of Learnt Embeddings*

Figs. 4 & 5 illustrate the model's effectiveness in learning the embedding space of Bitcoin transaction outpoints using the triplet loss framework. The first plot in Fig. 4 (left) displays the

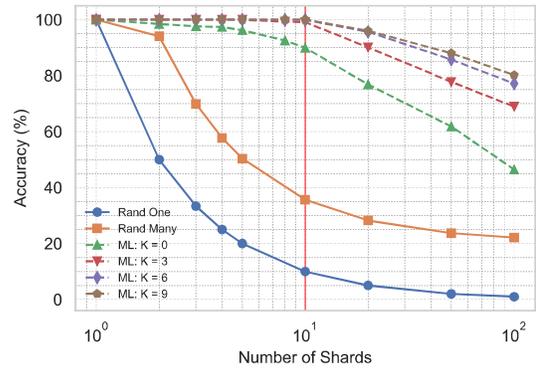

Fig. 3 ML significantly reduces parent-child shard allocation misassignment seen with random allocation.

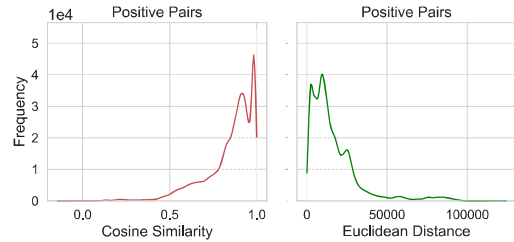

Fig. 4 Distribution of distances between positive pairs.

distribution of cosine similarities between anchor-positive pairs, representing parent-child relationships in Bitcoin transactions. The distribution is heavily skewed toward 1, peaking near 0.9–1.0, indicating successful representation of related transaction outpoints with highly similar embeddings. The narrow peak suggests that most positive pairs are closely aligned, characteristic of a well-learned embedding. The second plot (right) depicts the distribution of Euclidean distances for the same positive pairs, with a peak near 0 and a strong cluster of distances below 50,000. This aligns with the objective of minimizing distances between related transaction outpoints, as stated in Equation (3). The narrow spread further confirms that the model effectively clusters related transaction embeddings. Fig. 5 presents cosine similarity distributions between randomly selected anchors and embeddings that do not correspond to their positive. Each plot represents a different anchor, showing significant variance in similarity levels between anchors and non-positive embeddings. Some anchors, like Anchor 4, exhibit distinct distributions with a peak around -0.5, indicating successful separation from unrelated embeddings. In contrast, others, such as Anchor 2, display more widespread distributions with less separation, suggesting overlap between embeddings for unrelated transaction outpoints. Range from strongly negative to positive values demonstrates the model's ability to separate many non-related transactions from the anchors. However, the variability in peaks across different anchors suggests that some anchors have better-defined representations than others. While Fig. 5 demonstrates clear progress in clustering related transaction outpoints and separating unrelated ones, the variability seen suggests that refining the embedding space's structure may enhance accuracy through more consistent separation.

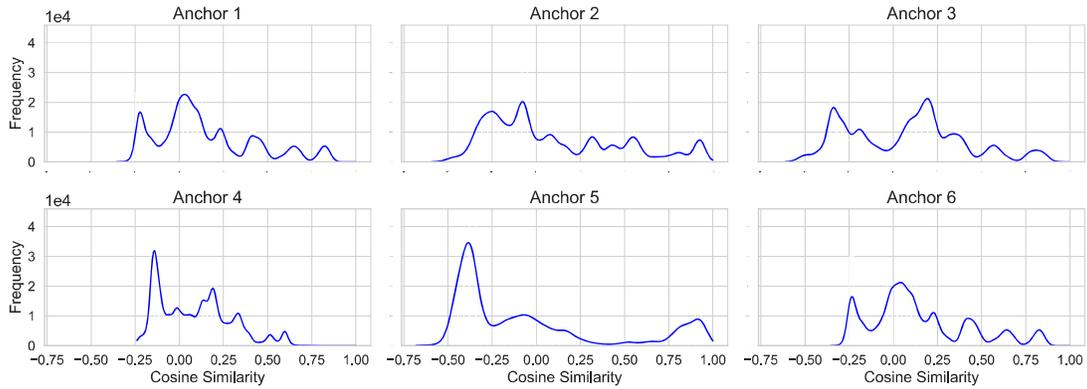

Fig. 5 Distribution of cosine similarity between six randomly chosen anchors compared to embeddings that are not their corresponding positive.

## V. CONCLUSION

We propose a ML framework to improve UTXO-based blockchain scalability through transaction sharding. By clustering transactions based on spending relationships using contrastive learning, our method assigns incoming transactions to shards likely to contain parent UTXOs, reducing cross-shard transactions and enhancing validation efficiency. Leveraging triplet loss and semi-hard negative sampling, our ML models effectively organize transaction outpoints by parent-child relationships using only local transaction features. This approach significantly improves shard allocation accuracy over random methods, demonstrating strong clustering of related transactions, though further refinement could improve separation of unrelated pairs. Future work will enhance consistency with advanced models and explore additional applications such as spending behavior classification and load balancing, with live blockchain testing to validate its utility.